\documentclass[fleqn,10pt]{wlscirep}
\usepackage[utf8]{inputenc}
\usepackage[T1]{fontenc}
\usepackage{graphicx}
\usepackage{dcolumn}
\usepackage{bm}
\usepackage{xcolor}
\usepackage{booktabs} 
\usepackage{enumitem}
\usepackage{amsmath}
\usepackage{multirow}
\usepackage{footnote}
\usepackage{tablefootnote}

\usepackage{graphbox}

\usepackage[ruled,algo2e]{algorithm2e}
\title{Constructing Energy-efficient Mixed-precision Neural Networks through Principal Component Analysis for Edge Intelligence}

\author[1,*]{Indranil Chakraborty}
\author[1]{Deboleena Roy}
\author[1]{Isha Garg}
\author[1]{Aayush Ankit}
\author[1]{Kaushik Roy}
\affil[1]{School of Electrical and Computer Engineering, Purdue University, West Lafayette, IN-47906, USA}

\affil[*]{ichakra@purdue.edu}

\begin{abstract}
The ‘Internet of Things’ has brought increased demand for AI-based edge computing in applications ranging from healthcare monitoring systems to autonomous vehicles. Quantization is a powerful tool to address the growing computational cost of such applications, and yields significant compression over full-precision networks. However, quantization can result in substantial loss of performance for complex image classification tasks. To address this, we propose a Principal Component Analysis (PCA) driven methodology to identify the important layers of a binary network, and design mixed-precision networks. The proposed Hybrid-Net achieves a more than 10\% improvement in classification accuracy over binary networks such as XNOR-Net for ResNet and VGG architectures on CIFAR-100 and ImageNet datasets while still achieving up to 94\% of the energy-efficiency of XNOR-Nets. This work furthers the feasibility of using highly compressed neural networks for energy-efficient neural computing
in edge devices.
\end{abstract}
\begin{document}

\flushbottom
\maketitle
%
%
\thispagestyle{empty}

\noindent Please note: Abbreviations should be introduced at the first mention in the main text – no abbreviations lists. Suggested structure of main text (not enforced) is provided below.

\section*{Introduction}

\section{Introduction}

The recent advent of `Internet of Things' (IOT) has deeply impacted our lives by enabling connectivity, communication and autonomous intelligence. With rapid proliferation of connected devices, the amount of data that needs to be processed is ever increasing. These data collected from numerous distributed devices are usually noisy, unstructured and heterogeneous \cite{Gubbi_2013}. Deep learning succeeds in reliably processing such complex and large volumes of data where conventional machine learning techniques fail \cite{Yao_2017}. Thus, it has become the driving force behind ubiquitous Artificial Intelligence (AI), and we see the pervasiveness of deep learning in various applications such as speech recognition, predictive systems and image and video classification \cite{krizhevsky2012imagenet,szegedy2015going,he2016deep,girshick2015fast}. 

Traditionally, IOT devices act as data collecting interfaces that feed the deep learning models deployed in centralized cloud computing systems. However, such systems have their own issues and vulnerabilities. In real-time application such as self-driving cars, the latency of communication between IOT devices and the cloud can pose a serious safety risk. As more IOT devices connect to the cloud, it strains the available shared bandwidth for communication. Furthermore, rising concerns around data privacy and over-centralization of information has propelled the need for decentralized user-specific systems \cite{kaufman2009data, gonzalez2012quantitative}. Edge computing \cite{li2016deepcham} is a promising alternative that enables IOT devices to process data, thus reducing communication overhead and latency and ensuring decentralization of data. The facilitation of on-chip analytics offered by edge computing can prove to be pivotal for autonomous platforms such as drones and self-driving cars as well as smart appliances. In addition, smart edge devices can play a significant role in healthcare monitoring systems and medical applications. Intelligent edge devices can be further leveraged for swarm intelligence based applications. However, computing in these resource constrained edge devices comes with its own challenges. Deep learning models are usually large in size and computationally intensive, thus making them difficult to implement in low-power and memory-constrained IOT devices. Thus, there is a need to design deep learning models which can perform effectively while requiring less memory and less computations.

One approach toward compressing neural network models is to modify the network architecture itself, such that it has fewer parameters, such as SqueezeNet \cite{iandola2016squeezenet}. Another method of compression is pruning which aims to reduce redundancies in over-parameterized networks. To that effect, researchers have investigated several network pruning techniques, both during training \cite{alvarez2017compression, weigend1991generalization} and inference \cite{han2015learning, ullrich2017soft}. 

A different technique of model compression is representing weights and activations with reduced precision. Quantized networks \cite{hubara2017quantized} help achieve reduction in energy consumption as well as improve memory compression compared to full-precision networks. Binary neural networks \cite{courbariaux2016binarized} are an extreme case of quantization where the activations and weights are reduced to binary representations. These networks drastically reduce the energy consumption by replacing the expensive multiply and accumulate (MAC) operations with simple add or subtract operations. This massive reduction in memory usage and computational cost make them particularly suitable for edge computing. However, despite these benefits, the networks suffer from performance and scalability issues, especially for complex pattern recognition tasks. Several training algorithms \cite{rastegari2016xnor} have been proposed to optimize network performance to achieve state-of-art accuracy in extremely quantized neural networks. Although such training methodologies recover the performance hit caused by binarizations weights alone, they fail to completely counter the degradation caused by binarizing both weighs and activations. 

In this work, we present Hybrid-Net, a mixed-precision network topology fashioned by the combination of binary and high-precision inputs and activations in different layers of a network. We use Principal Component Analysis (PCA) to determine significance of layers based on the ability of a layer to expand data into higher dimensional space, with the ultimate aim of linear separability. Viewing a neural network as an iterative projection of input onto a successively higher dimensional manifold at each layer, until the data is eventually linearly separable allows us to identify layers that contribute relevant transformations. Following the algorithm in \cite{garg2018low}, we find the `significant dimensions' in a layer as the number of dimensions that cumulatively explain 99\% of the total variance of the output activation map generated by that layer. Since we want the data to be expanded into higher dimensions at each layer, we deem the layers at which significant dimensions increase from the previous layer as significant.  Following the identification of significant layers, we increase the bit-precision of the inputs and weights of those layers, keeping the rest of the layers entirely binary. Traditionally, PCA has been used primarily as a dimensionality reduction technique. It was also recently used to identify redundancies in different layers of a neural network and prune out the redundant features\cite{garg2018low}. We propose a methodology where we use PCA in a reverse manner, i.e., to increase the precision of the important layers. Hybrid-Net remarkably improves the performance of extremely quantized neural networks, while keeping the activations and weights of the most of the layers binary. This ensures low energy consumption and high memory compression of extremely quantized neural networks while achieving significantly enhanced classification performance compared to binary networks such as XNOR networks. This work not only achieves signficant progress in the challenge of quantizing neural networks to binary representations but also paves way for optimized yet highly accurate quantized networks suitable for enabling intelligence at the edge. 
\section{Related Work}
Various techniques have been proposed to improve the performance of quantized networks. Fully binary networks \cite{courbariaux2016binarized, rastegari2016xnor} are constructed by replacing the activations with their sign. However, these networks usually suffer from significant degradation in accuracy, especially for larger datasets such as CIFAR-100 and ImageNet. One intuitive way of recovering quantization errors is using wider networks \cite{mishra2017wrpn} but it comes at the cost of increased energy consumption. There have been efforts focusing on gradient calculations for approximated sign functions to ameliorate the effect of binarization \cite{liu2018bi}. More general quantization schemes have also been explored for weights and activations \cite{zhou2016dorefa, zhou2017balanced}. Although weight quantization can be compensated by training the network with quantized weights \cite{hubara2017quantized}, it has been observed that input quantization pose a serious challenge to classification performance for precisions lower than 4 bits. One approach that addresses this challenge involve clipping the activations by setting an upper bound. Although this approach seems to be heuristic, recent efforts have focused on using trainable quantization that can be dynamically manipulated\cite{zhang2018lq,jung2018joint}. One such approach involves parameterized clipping where the clipping level is dynamically adjusted through gradient descent \cite{choi2018pact}. Another approach proposed the use of batch-normalization layers after ReLU activations to bound the activation values for effective quantization \cite{graham2017low}. Note, that most of these works focus on optimizing the activations when the quantization precision is 2 bits or more. Binary networks with both 1-bit activations and weights, despite offering the most benefits in terms of computation cost and memory compression, still suffer from significant degradation in performance compared to full-precision networks. 

An alternative path towards improving the accuracy of binary neural networks focuses on network design techniques. To that effect, improved input representations through shortcut connections in deep networks can significantly improve performance of binary neural networks without any increase in computation cost \cite{liu2018bi}. This is because shortcut connections are usually identity in nature and do not comprise of expensive MAC operations. Combinations of different kinds of input precisions have also been explored across different layers to circumvent the significant decrease in classification accuracy of such binarized networks \cite{prabhu2018hybrid}. There has been considerable effort in making the search for optimum neural architecture more sophisticated through efficient design space exploration \cite{wu2018mixed}. A theoretical approach towards predicting layer-wise precision requirement has been also explored  \cite{sakr2018per}. 
Our work differs from most of the current efforts in quantized neural networks as it lies in the realm of hybrid network design for more optimal performance of neural networks where most of the layers still have 1-bit weights and activations. This motivates us to propose an algorithm to identify important layers and judiciously reinforce those particular layers with higher bit-precision representation. To follow such a motivation, it is necessary to understand the significance of layers, which we explain in the next section.
\begin{figure}[t]
	\centering
	\includegraphics[width=4.9in, keepaspectratio]{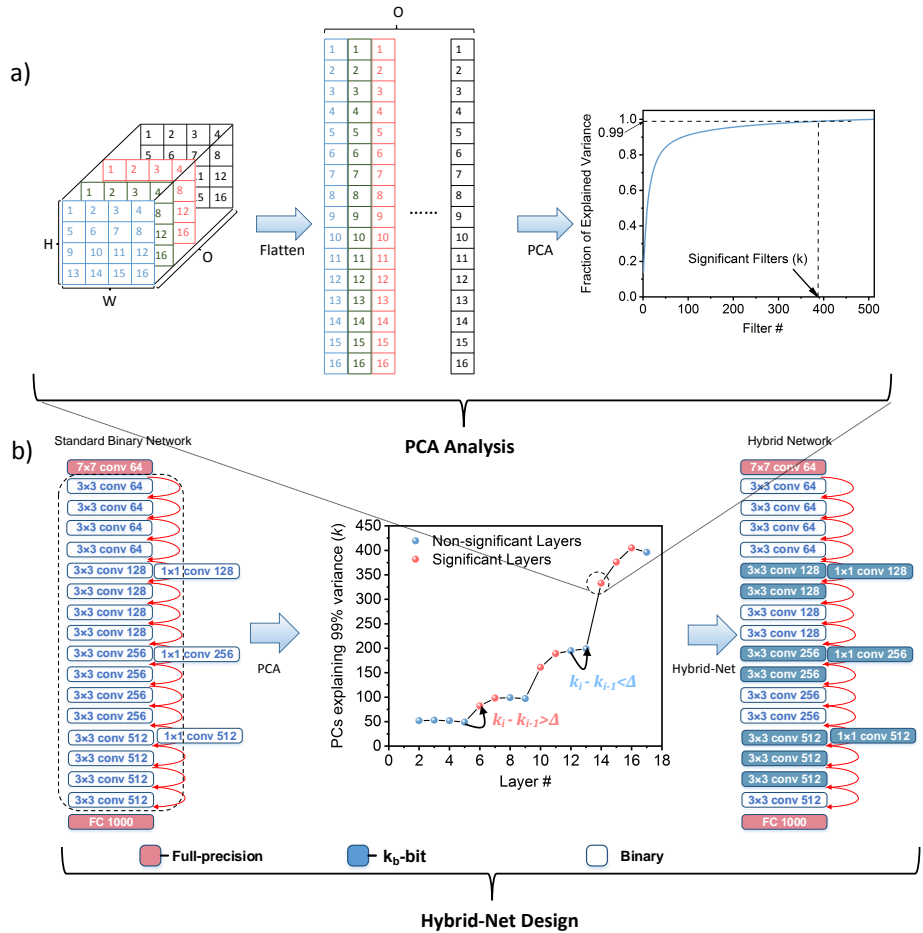}

	\caption{Illustration to show PCA analysis and subsequent Hybrid-Net design. (a) The PCA function in Algorithm. 1 for a particular layer showing flattening of an instance of a 4-D tensor and subsequent PCA analysis yielding the plot showing how the cumulative explained variance accrues with the number of filters in a particular layer. Number of Significant filters, $k$ is defined as the number of filters at which the cumulative sum reaches a threshold (say 0.99). (b) The Main() function in Algorithm. 1 is explained as we take a standard binary network (leftmost, with first and final layers having full-precision weights and activations) and perform the aforementioned PCA function on each binary layer (shown in white). The resulting plot (middle) shows the layer-wise variation in $k$ in a ResNet-18 on ImageNet for example. A layer is considered significant (shown by red markers), when $k$ increases by at least $\Delta$. The new Hybrid-Net (rightmost) is designed by increasing the precision of weights and activations for the significant layers (shown in blue).}
	\label{fig:pcaexp}
	\vspace{-4mm}
\end{figure}
\section{PCA-driven Hybrid-Net Design}
A Hybrid-Net is a neural network that employs two different bit precisions for its weights and activations. The base network is of low precision, for example 1 bit, and certain layers are selected and set to a higher bit precision. For selecting the layers, we use Principal Component Analysis (PCA) on the output feature maps of each of the layers. The input to any layer is binarized and convolved with the weight filters. We perform PCA on the resulting output tensor. This is performed individually on the output tensor of every layer. Given any set of correlated variables, such as the feature maps, PCA does an orthogonal transformation to map them to uncorrelated variables called Principal Components(PCs), which also form the orthogonal basis set for these tensors. Each of these resulting basis vectors identify directions of varying variance in the data, and are ordered in decreasing manner, with the first vector in the direction of highest variance. We perform such PCA on each convolutional layer in a standalone fashion based on the transformation that it applies on its input. It is applied only on the linear portion of the network before non-linear activation. The aim of PCA is to identify redundancy and how many filters in that layer are necessary to give us a near perfect reconstruction of the output. Based on the redundancy data obtained from the PCA, we define our own significance metric to obtain significant layers from the network. 

In a neural network, each layer applies a transformation on its input and projects it to a new feature space of ideally higher or same dimension with the objective of achieving linear separability. PCA provides the ability to study the directions of maximum variance in the input data. The pre-ReLU activation map generated by a filter is considered to composed of many instances of that particular filter. Performing PCA and finding the number of filters needed to explain a pre-defined cumulative percentage of variance identifies the number of significant dimensions for each layer. More the number of principal components needed to preserve a significant percentage, say $T$, of the total variance in the input, lesser is the redundant information carried by those tensors, and higher is the significant dimensionality of those tensors. Ideally, we want the number of PC's required to explain $T$\% of the total variance of the feature space to increase as we move deeper into the network in order to extract more uncorrelated, unique features from the data, and project it into a higher dimensional space that will eventually lead to linear separability at the classifier layer. Thus, the layers for which the number of PCs explaining variance in the output data is more than that in the input data, contribute to significant transformations on the input data. Note, the significant dimensionality of layers does not always monotonically increase. However, regardless of the trend, it provides us a way of judging the pliability of a layer to binarization. In this section, we propose a methodology to identify these significant layers and subsequently design mixed-precision networks by increasing the bit-precision of those layers. 

\begin{algorithm2e}[h!]
\SetAlgoLined

 \SetKwFunction{FPCA}{PCA}
 \SetKwProg{Fn}{Function}{}{}
 \Fn{\FPCA{activations,layer,T}}{
    \nl [M,H,W,O] $\longleftarrow$ size(activations[layer])\;
    \nl \textit{act\_fl} $\longleftarrow$ flatten(activations[layer],M*H*W,O)\;
    \nl run PCA on \textit{act\_{fl}}\;
    \nl \textit{tot\_{var}} $\longleftarrow$ total variance\;
    \nl \textit{cum\_{var}} $\longleftarrow$ cumulative sum of variance\;
    \nl k $\longleftarrow$ num of components with \textit{cum\_{var}}<T*\textit{tot\_{var}}\;
    \nl \KwRet k\;}{}
  \SetKwFunction{Fmain}{Main}
  \SetKwProg{Fn}{Function}{}{}
   \Fn{\Fmain{}}{
   \nl  Train a N-layer binary network\;
   \nl Set Threshold $T$ \;
   \nl Set Delta $\Delta$ \;
   \nl Set Delta $k_b$ \tcp*{Bit Precision of significant layers}
   \nl \For{$i\gets0$ \KwTo $N-1$}{
        \nl $k[i] \longleftarrow PCA(activations, i, T)$\; 
        \If{$k[i]-k[i-1]>\Delta$ \& $i>0$}{
        \nl Add to Significant Layer list\;
        }
        
    }
    \DontPrintSemicolon
    \nl Create Hybrid-Net: \;
    \For{$i\gets1$ \KwTo $N-1$}{
    \uIf{$i \in$ $Sig_{layer}$}{
        $Prec-{layer} \longleftarrow k_b$\;
        }
    \Else{
        $Prec-{layer} \longleftarrow 1$\;
        
    }
    }
    \nl Initialize weights with same seed\;
    \nl Train Hybrid-Net\;

   }{}

 \caption{Hybrid-Net Design Methodology}
\end{algorithm2e}
\subsection{PCA-driven identification of significant layers }

We perform our analysis on activations of each layer, which provide a notion of activity of each filter in that layer. Let us consider the activation matrix of the $L^{th}$ layer, $X_L$. Layer $L$ has $O$ filter banks, each containing $I$ filters of size $k\times k$. $I$ and $O$ are the number of input and output channels. The first element of the $i^{th}$ output map of $X_L$ is the result of convolution of the first $k\times k \times I$ sized input patch with the $i^{th}$ filter bank. The rest of the elements of any particular output map of $X_L$ can be obtained by striding over the entire input. Thus, if we consider $M$ as the size of a minibatch, $X_L$ is a 4-dimensional tensor of size $M\times H\times W\times O$ where $H$ and $W$ are the height and width of the each output map. If we flatten the 4-D data $X_L \in \mathbb{R}^{M\times H\times W\times O}$ into a 2-D matrix $Y_L \in \mathbb{R}^{M*H*W\times O}$, we would obtain $M*H*W$ samples, each containing $O$ elements equivalent to the number of filter banks. This process is shown in Fig. \ref{fig:pcaexp} (a).

When PCA is performed over the aforementioned 2-D matrix $Y_L$, the singular value  decomposition (SVD) of the mean normalized, symmetric matrix $Y_L^TY_L$ generates $O$ eigenvectors $v_i$ and eigenvalues $\lambda_i$. The total variance in the data is given by sum of the variances of individual parameters:
\begin{equation}
    Var = \sum_i^O(\sigma_{ii}^2) = Tr(Y_L^TY_L)
\end{equation}
The contribution of any component, $\lambda_i$, towards the total variance can be expressed as $\lambda_i/Var$. To calculate the number of significant components, we set a threshold value $T$ which is amount of variance the first $k$ significant components are able to explain. This can be expressed as:
\begin{equation}
    \frac{\sum_{i=1}^{k}(\lambda_{i}^2)}{\sum_{i=1}^O(\lambda_{i}^2)} = T
\end{equation}
An example of a typical curve of the cumulative sum of variance for different filter numbers, obtained by PCA, is shown in Fig. \ref{fig:pcaexp} (a) (rightmost). As the PCA analysis produces the $k$ most significant components to explain $T$ fraction of the total variance, we proceed to identify the significant layers. We define a significant layer as the layer which transforms the input data such that the number of significant components to explain $T$ fraction of the variance, increase from that required for the output of previous layer. Let $k_i$ be the number of significant components corresponding to the $i^{th}$ layer. Then, it can be said that layer $i$ contributes a relevant transformation on the input data if $k_{i} >k_{i-1}$. It means that the layer requires more significant components to explain the variance in the data at its output than the previous layer. However, for a better control on deciding the important layers, we check the condition whether $k_{i}-k_{i-1}>\Delta$ to determine if the $i^{th}$ layer is significant. This is explained in Fig. \ref{fig:pcaexp} (b) (middle) where the dots marked in red denote the significant layers where $k_{i}-k_{i-1}>\Delta$. 
\subsection{Hybrid-Net Design}

The PCA analysis helps us identify a set of important layers in an N-layer network. We design a hybrid network, `Hybrid-Net', where we set the bit-precision of weights and inputs of the important layers to a higher value, $k_b = 2,4$, than the other layers which have binary weights and inputs. This is shown in Fig. \ref{fig:pcaexp} (b) (rightmost). The weights and inputs of the first and the final layers of a N-layer network are kept full-precision, according to standard practice \cite{rastegari2016xnor,zhou2016dorefa,choi2018pact}. The quantization algorithm for any $k_b$-bit quantized layer can be readily derived from XNOR-Net \cite{rastegari2016xnor} where the quantized levels are: 
\begin{equation}
\centering
    q_{k-b}(x) = 2(\frac{\lfloor{(2^{k_b}-1)(x+1)/2}\rfloor}{2^{k_b}-1}-\frac{1}{2})
\end{equation}
In a layer with $k_b$-bit weights and activations, $q_{k_b} (x)$ is used instead of the $sign$ function in layers with binary weights and activations. We use a slightly modified version of quantized networks, proposed in \cite{rastegari2016xnor}, where the weights have a scaling factor $\alpha $ instead of just being quantized. The convolution operation in between inputs $X$ and weights $W$ in such a network is approximated as:
\begin{flalign}
\text{Binary:}  && X*W &\approx (sign(X)*sign(W))\odot\alpha &\\ \text{$k_b$-bit:} && X*W &\approx (q_{k-b}(X)*q_{k-b}(W))\odot\alpha
\end{flalign}
Here, $\alpha$ is the L1-norm of $W$ and act as a scaling factor for the binary weights. In binary layers, the activation gradients are clipped such that they lie between -1 and 1. In the $k_b$-bit layers, we get rid of the activation gradient clipping for better representation. Each layer of a N-layer Hybrid-Net have either binary or $k_b$-bit weight kernels and the activations after each convolution are again quantized before passing to the next layer. 

Hybrid-Net is expected to have a higher computation cost than a binary network. The parameter $\Delta$ decides the number of important layers to consider and hence a penalty is incurred due to increase in bit-precision. We can estimate the penalty in computation cost incurred due to the increase in bit-precision a in network with $L_S$ number of significant layers as
\begin{equation}
    Penalty = \frac{\sum\limits_{i\notin Sig_{Layer}}{B_i}+\sum\limits_{i\in Sig_{Layer}}{B_i\times p}}{\sum\limits_{i=1}^{N}{B_i}}
\end{equation}
Here $B_i$ is the computation cost of a binary layer and $p$ is the overhead of $k_b$-bit computation over binary computation. We will present a detailed analysis of energy consumption and memory usage later in the manuscript. 

For residual networks, ResNets, we include another design feature in addition to the PCA-driven Hybrid-Net, improving input representations through residual connections. This has been alluded to by Liu et al in \cite{liu2018bi} where adding identity shortcut connections at every layer improves representational capability of binary networks. In standard residual networks \cite{he2016deep}, such identity connections are added to address the vanishing gradient problem in deep neural networks. However, in case of binary networks, these connections serve to provide an improved representation by carrying floating-point information from the previous layer. As a result, the Hybrid-Net design also considers the effect of adding such highway connections at every layer. Note, in case of convolution layers which induce a change in size of each feature map, the shortcut connections consist of $1\times 1$ convolution weight layers to account for the change in size \cite{he2016deep}. 

\begin{figure}[t]
\centering
    \includegraphics[align=c,width=\textwidth]{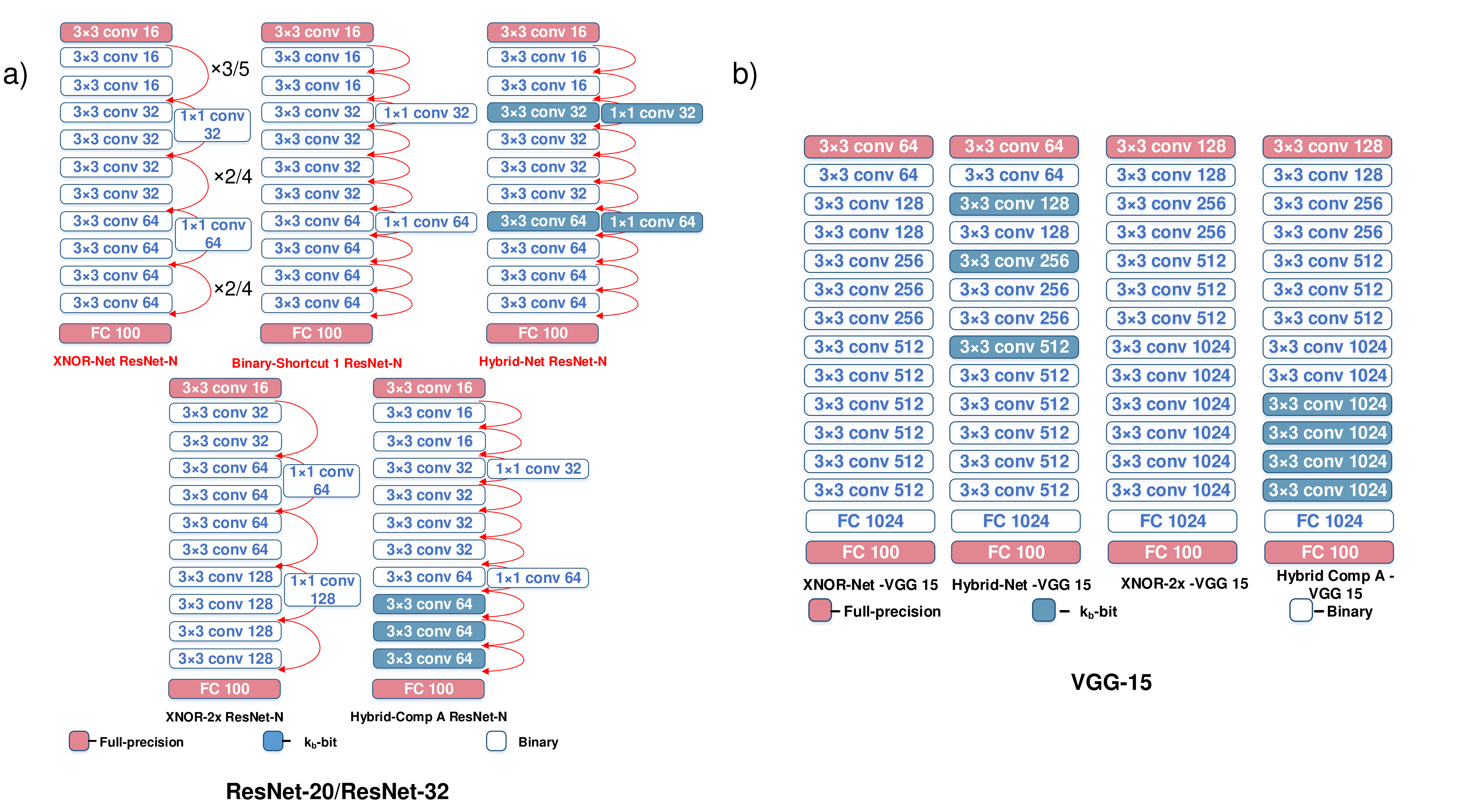}

  \caption{Different network configurations based on a) ResNet-20/ResNet-32 and b) VGG-15 architectures showing binary, $k_b$-bit and full-precision layers in each of the networks as described in Section 4.2. The width of the layer shown in this figure is for CIFAR-100 dataset}
  \label{fig:resnetvgg}
\end{figure}

\section{Experiments, Results and Discussion}

\subsection{Experiments}
We evaluated the performance of all the networks described in this section 4.2 in PyTorch \cite{paszke2017automatic}. We perform image classification on the datasets CIFAR-100 \cite{krizhevsky2009learning} and ImageNet \cite{deng2009imagenet}. The CIFAR-100 dataset has 50000 training images and 10000 testing images of size $32\times32$ for 100 classes. For the CIFAR-100 dataset, we explore the proposed Hybrid-Net design methodology on standard network architectures, ResNet-20, ResNet-32 and VGG-15, where the training algorithm for the quantized layers has been adopted from \cite{rastegari2016xnor}. We extended our analysis to the ImageNet dataset \cite{deng2009imagenet} which is the most challenging dataset pertaining to image classification tasks. It consists of 1.2 million training images and 50000 validation images divided into 1000 categories. For simplicity, we considered ResNet-18 for our ImageNet evaluation. To compare against the proposed Hybrid-Net designs, we explore different network configurations as baselines, for ResNet, shown in Fig. \ref{fig:resnetvgg} (a) and VGG architectures, shown in Fig. \ref{fig:resnetvgg} (b). XNOR-Net is the base skeleton network we design other networks on. Binary-Shortcut is same as XNOR-Net except it has residual connections every layer similar to \cite{liu2018bi}. Hybrid-Comp A is formed by inter-layer sectioning, i.e., dividing the network into 2 parts ($N-k$ binary and $k$ $k_b$-bit layers) where $N$ is the number of the layers between the first and last layer. The widths of the network architectures, shown in Fig. \ref{fig:resnetvgg} (a) and Fig. \ref{fig:resnetvgg} (b) are for CIFAR-100 dataset. For ImageNet, we have used a wider network architecture, which we describe in Table. \ref{imtable}. We have also compared our proposed Hybrid-Net designs against state-of-art quantized networks such as \cite{choi2018pact, zhou2016dorefa, zhang2018lq} for ImageNet. We performed simulations for 5 different random initializations for all networks on CIFAR-100 dataset. For simplicity, we performed simulations for 3 different random initializations on 4 selected networks and 1 initialization for the rest of the networks on ImageNet dataset. The accuracy reported for both datasets is the top-1 accuracy and the Mean  $\pm$  SD accuracy provides the mean and standard deviation of accuracies obtained for different random initializations.
\begin{table}[t]
\centering
	\caption{Networks architectures for ImageNet classification task}
	\label{imtable}
\begin{tabular}{|c|}
\hline
\textbf{ResNet - 18}                         \\ \hline      
7$\times$7 conv 64 stride 2          \\ \hline     
3$\times$3 maxpool stride 2          \\ \hline     
3$\times$3 conv 64 stride 1 ($\times$ 4)\\ \hline  
3$\times$3 conv 128 stride 2             \\ \hline 
3$\times$3 conv 128 stride 1 ($\times$ 3)\\ \hline 
3$\times$3 conv 256 stride 2            \\ \hline  
3$\times$3 conv 256 stride 1 ($\times$ 3)\\ \hline 
3$\times$3 conv 512 stride 2             \\ \hline 
3$\times$3 conv 512 stride 1 ($\times$ 3)\\ \hline 
Linear 1000                             \\ \hline  
\end{tabular}
\end{table}
\subsubsection{Energy efficiency and Memory compression}
We have briefly alluded to the possible penalty incurred due to increasing the bit-precision of certain layers in a network. To identify its effect with respect to the entire network metrics and further illustrate the benefits of the proposed Hybrid-Nets, we perform a storage and computation analysis to calculate the energy efficiency and memory compression of the proposed networks. For any two networks A and B, the energy efficiency and memory compression of Network A with respect to Network B can be defined as: 
\begin{equation}
\begin{aligned}
    \text{Energy Efficiency } (E.E) &= \frac{E_A}{E_B} \\
    \text{Memory Compression } (M.C) &= \frac{M_A}{M_B} 
\end{aligned}
\end{equation}
where $E_A$ and $E_B$ are the energy consumed by Network A and Network B respectively, $M_A$ and $M_B$ are the memory used for storing the weights of Network A and Network B, respectively. We estimate energy efficiency (E.E) and memory-compression (M.C) with respect to an full-precision network and normalize it with respect to an XNOR-Net network which is an entirely binary network except the first and final layer. Thus, the normalized E.E ($E.E_{Norm}$) and normalized M.C ($M.C_{Norm}$) of any network A can be written as:
\begin{align}
\begin{split}
    E.E (A) &= \frac{\sum_i{E_i(FP)}}{\sum_i{E_i(A)}} \\
    E.E_{Norm} (A) &= \frac{E.E (A)}{E.E (XNOR)} \\
    M.C (A) &= \frac{\sum_i{M_i(FP)}}{\sum_i{M_i(A)}} \\
    M.C_{Norm} (A) &= \frac{M.C (A)}{M.C (XNOR)}
    \end{split}
\end{align}
Here, $E_i(FP) (M_i(FP))$ is the energy (memory) consumed by the $i^{th}$ layer of a network with full-precision weights and activations whereas $E_i(A) (M_i(A)$ is the energy (memory) consumed by the $i^{th}$ layer of any network A under consideration.

\subsection{Results - PCA}
We perform PCA analysis on the activations of each convolutional layer and extract the number of principal components required to explain a $T$ fraction of variance in the data. The design parameters such as $T$ and subsequently $\Delta$ are heuristically. For all analysis, we fix $T=99\%$ as this makes the increases in significant components, $k$ across various layers clearly distinguishable. The $\Delta$ values are chosen based on the variation in $k$ across layers. A higher $\Delta$ value yields less number of significant layers. For clarity, we perform our analysis for various $\Delta$ values.
\subsubsection{ResNet Architectures - CIFAR-100}
\begin{figure*}[]
	\centering
	\includegraphics[width=4.5in, keepaspectratio]{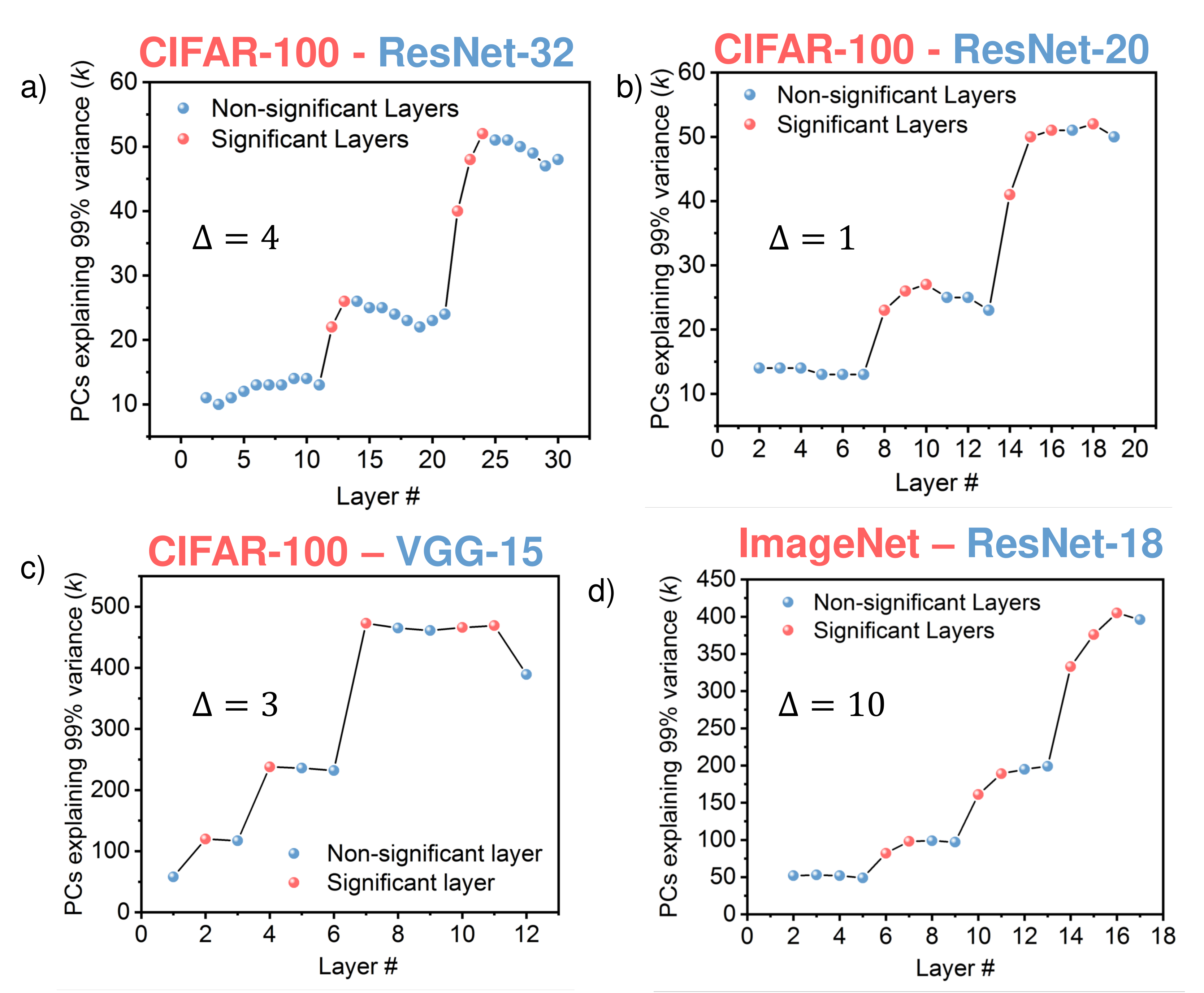}

	\caption{PCA analysis plot showing the number of principal components required to explain 99\% variance at the output of convolutional layers across different layers for a) ResNet-32 ($\Delta=4$), b) ResNet-20 ($\Delta=1$), and c) VGG-15 ($\Delta=3$) on CIFAR-100 dataset and d) for ResNet-18 ($\Delta=30$) on ImageNet dataset}
	\label{fig:pca}

\end{figure*}

For ResNet architectures, we perform the PCA on a plain version of a binary network devoid of any residual connections. We decided to do this to isolate the effect of the convolution layers on the activations, instead of having residual connections. This is done because we focus on the quantization of the filters of the layers and the residual additions may distort the output feature space and hence the information we seek from it. Fig. \ref{fig:pca} (a) and (b) shows the variation in the number of filters required to explain $T = 99\%$ with different layers for ResNet-20 and ResNet-32 architectures respectively. As expected, the maximum change in $k$ occurs when the number of output channels increase. However, we observe a trend in both networks, that the layers just after the output channels increase from, say 16 to 32 or 32 to 64, attribute for the maximum change in the number of significant filters. Based on our criteria for significant layers, discussed in Section 3.1, we fix a $\Delta = 1$ for Resnet-20 and $\Delta = 4$ for ResNet-32 to identify the layers where the number of significant components undergo a change more than $\Delta$. Fig. \ref{pca} (a) also shows those layers marked by red dots. Note, by varying the $\Delta$, more or less number of layers can be considered as significant. After performing this analysis on a plain version of the ResNet architecture, we perform network simulations on the standard version with residual connections. 
\begin{table}[h!]
\centering
\caption{Significant Layers identified by PCA analysis}
	\label{pca}
\begin{tabular}{|l|l|}
\hline
\multicolumn{2}{|c|}{\textbf{CIFAR-100}} \\
\hline
Network Arch           & Significant layers       \\ \hline
ResNet-20 ($\Delta=1$) & 8, 9, 10, 14, 15, 16, 18 \\ \hline
ResNet-20 ($\Delta=2$) & 8, 9, 14, 15             \\ \hline
ResNet-32 ($\Delta=4$) & 12, 13, 22, 23, 24       \\ \hline
VGG-15 ($\Delta=3$)    & 3, 5, 8, 11, 12          \\ \hline
VGG-15 ($\Delta=10$)   & 3, 5, 8                  \\ \hline
\hline
\multicolumn{2}{|c|}{\textbf{ImageNet}}\\
\hline
Network Arch           & Significant layers       \\ \hline
ResNet-18 ($\Delta=30$) &  6, 10, 14, 15\\ \hline
ResNet-18 ($\Delta=20$) &  6, 10, 11, 14, 15, 16   \\ \hline
ResNet-18 ($\Delta=10$) &  6, 7, 10, 11, 14, 15, 16   \\ \hline

\end{tabular}
\end{table}
\subsubsection{VGG Architectures - CIFAR-100}
For VGG architecture, we perform the PCA of a binary network which has binary weights and activations for all layers except the first and the last. Fig. \ref{fig:pca} (c) shows the plot showing how the number of filters required to explain $T = 99\%$ of the variance changes with different layers for a VGG-15 architecture. We observe that the number of significant filters mostly increase when the number of filter bank increases at a particular layer. For rest of the layers, it remains fairly constant. As the PCA plot shows very little variation across layers, we consider a relatively lower $\Delta = 3$ with respect to the number of filters. We mark the significant layers by red dots in Fig. \ref{fig:pca} (c). Table. \ref{pca} lists the different combination of significant layers obtained from ResNet and VGG architectures through the PCA analysis for different $\Delta$ values for CIFAR-100 dataset. Note, we did not choose a lower $\Delta$ for ResNet-32 as it would have included many layers which would increase the computation cost without a significant benefit in accuracy.

\subsubsection{ResNet Architectures - ImageNet}
We further perform PCA analysis on ResNet-18 architecture for the ImageNet dataset. Fig. \ref{fig:pca} (d) shows the plot showing how the number of filters required to explain $T = 99\%$ of the variance changes with different layers for ResNet-18 for $\Delta=10$. The significant layers identified by our proposed methodology are marked with red dots. We observe a similar trend as in case of CIFAR-100, that maximum increase in the number of significant filters, $k$, occur in the first few layers after every change in filter size. We perform the PCA analysis for $\Delta=10, 20, 30$ to identify the significant layers, listed in Table. \ref{pca}.

\subsection{Image Classification Results - CIFAR-100}

\subsubsection{ResNet Architectures}
The ResNet-N architecture consists of N-1 convolution layers and a fully-connected classifier. As discussed before, the first convolution layer and the classifier have full precision inputs and weights. For CIFAR-100 dataset, we consider $N=20$ and $N=32$. Further, we consider a slightly modified version of ResNet, where we add identity shortcut connections at every layer instead of every two layers for better input representation, as discussed earlier. We increase the bit-precision of weights and inputs of the layers obtained from PCA analysis to bit precisions $k_b=2$ and $k_b=4$ to form Hybrid-Net ($k_b$, $k_b$). The rest of the layers have binary representations for weights and inputs. We also compare the proposed Hybrid-Net with Hybrid-Comp A ($k_b$, $k_b$) ($k$), which is formed by splitting the entire network into $N-k$ binary and $k$ $k_b$-bit sections. Table. \ref{results} shows that accuracy, energy efficiency and memory compression of the proposed Hybrid-Net based on ResNet-20 and ResNet-32 in comparison to XNOR-Net and other kinds of hybrid networks discussed in Fig. \ref{fig:resnetvgg}. 
\begin{table}[h!]
\centering
\caption{Comparison of different networks on CIFAR-100}
\label{results}
\begin{tabular}{|l|c|c|c|c|}

\hline
\multicolumn{5}{|c|}{\textbf{ResNet-20}}                                                                                   \\ \hline
\multicolumn{5}{|c|}{\textbf{FP Accuracy - 69.49\%}}                                                                                   \\ \hline
\multicolumn{5}{|c|}{\textbf{$E.E (XNOR)$ - 16.35}, \textbf{$M.C (XNOR)$ - 17.26}}  \\ \hline
\textit{Network Type}                                                  & \multicolumn{1}{l|}{\textit{Best Accuracy (\%)}} & \multicolumn{1}{l|}{\textit{Mean   $\pm$  SD Accuracy (\%)}} & \multicolumn{1}{l|}{\textit{$E.E_{Norm}$}} & \multicolumn{1}{l|}{\textit{$M.C_{Norm}$}} \\ \hline
XNOR                                                                   & 50.50                                            & 50.23   $\pm$  0.21                                                 &  1                                          &   1                                         \\ \hline
Binary- Shortcut 1                                                              & 54.16                                            & 53.92   $\pm$  0.21                                                 & 0.99                                           &      1                                      \\ \hline
Hybrid-Net (2,2) ($\Delta$=1)                                             & 62.84                                            & 62.5  $\pm $ 0.24                                                  &      0.87                                      &          0.77                                  \\ \hline
Hybrid-Net (2,2) ($\Delta$=2)                                             & 60.93                                            & 60.53   $\pm$  0.39                                                 &     0.93                                       &    0.88                                        \\ \hline

\begin{tabular}[c]{@{}l@{}}Hybrid-Net (4,4)  ($\Delta$=1)\end{tabular} & 63.88                                            & 63.38   $\pm$  0.49                                                 &    0.7                                        &  0.53                                          \\ \hline
\begin{tabular}[c]{@{}l@{}}Hybrid-Net (4,4)   ($\Delta$=2)\end{tabular} & 61.62                                            & 61.53   $\pm$  0.1                                                  &    0.82                                        &   0.7                                          \\ \hline
Quantize(2,2)                                                          & 65.81                                            & 65.19   $\pm$  0.49                                                  & 0.73                                           & 0.65                                            \\ \hline
Hybrid-Comp A (2,2)(k=6)                                                     & 62.36                                            & 61.79  $\pm$  0.34                                                  &  0.88                                          &  0.71                                          \\ \hline

XNOR2x                                                                 & 63.03                                            & 62.81  $\pm$  0.14                                                 & 0.39                                           & 0.33                                            \\ \hline
\multicolumn{5}{|c|}{\textbf{Resnet-32}}       \\   \hline    \multicolumn{5}{|c|}{\textbf{FP Accuracy - 70.62\%}}                                                                                   \\ \hline
 \multicolumn{5}{|c|}{\textbf{$E.E (XNOR)$ - 18.42, $M.C_{Norm}$ - 20.44}}                                                                                   \\ \hline                                          
\textit{Network Type}                                                  & \multicolumn{1}{l|}{\textit{Best Accuracy (\%)}} & \multicolumn{1}{l|}{\textit{Mean  $\pm$  SD Accuracy (\%)}} & \multicolumn{1}{l|}{\textit{$E.E_{Norm}$}} & \multicolumn{1}{l|}{\textit{$M.C_{Norm}$}} \\ \hline
XNOR                                                                   & 53.89                                            & 53.48  $\pm$  0.27                                                 &     1                                       & 1                                           \\ \hline
Binary- Shortcut 1                                                             & 58.98                                            & 58.23  $\pm$  0.61                                                 &      0.99                                      &                                    1        \\ \hline
Hybrid-Net (2,2) ($\Delta$=4)                                             & 64.34                                         & 63.75  $\pm$  0.39                                                   &          0.94                                  &                                     0.87       \\ \hline
\begin{tabular}[c]{@{}l@{}}Hybrid-Net (4,4) ($\Delta$=4)\end{tabular} & 64.45                                            & 64.28  $\pm$  0.18                                                  &                      0.84                      & 0.69                                            \\ \hline
Quantize(2,2)                                                          & 68.04                                            & 67.73  $\pm$  0.21                                               &                  0.7                          & 0.61                                            \\ \hline
Hybrid-Comp A (2,2)                                                   & 62.41                                            & 62.15  $\pm$  0.2                                                &                     0.91                       &         0.76                                   \\ \hline

XNOR2x                                                                 & 65.20                                            & 65.11  $\pm$  0.07                                                  &                         0.38                   &                             0.31               \\ \hline
\multicolumn{5}{|c|}{\textbf{VGG-15}}   \\ \hline              \multicolumn{5}{|c|}{\textbf{FP Accuracy - 68.31\%}}                   \\ \hline
 \multicolumn{5}{|c|}{\textbf{$E.E (XNOR)$ - 21.77, $M.C_{Norm}$ - 26.24}}                                                                                                                                                                                                                         \\ \hline
\textit{Network Type}                                                  & \multicolumn{1}{l|}{\textit{Best Accuracy (\%)}} & \multicolumn{1}{l|}{\textit{Mean   $\pm$  SD Accuracy (\%)}} & \multicolumn{1}{l|}{\textit{$E.E_{Norm}$}} & \multicolumn{1}{l|}{\textit{$M.C_{Norm}$}} \\ \hline
\begin{tabular}[c]{@{}l@{}}
XNOR \\  \end{tabular}                                                                  &   54.30                                               &  54.23  $\pm$  0.1                                                     &   1                                         &                                   1         \\ \hline
Hybrid-Net (2,2) ($\Delta$=3)                                             & 61.81                                            & 61.67  $\pm$  0.08                                                  &                           0.84                 &                        0.75                    \\ \hline
Hybrid-Net (2,2) ($\Delta$=10)                                            &   60.13                                               &   59.87  $\pm$  0.25                                                    &    0.93                                        &                                    0.92        \\ \hline

Hybrid-Net (4,4)   ($\Delta$=3) & 63.38                                            & 63.12  $\pm$  0.15                                                  & 0.64                                           &                                 0.5           \\ \hline
Hybrid-Net (4,4) ($\Delta$=10)                                            &   60.37                                               &     60.06  $\pm$  0.18                                                  &                                0.81            &                                       0.8     \\ \hline
Quantize(2,2)                                                          &  68.90                                                &    68.63  $\pm$  0.28                                                  &             0.65                               &                                           0.55 \\ \hline
Hybrid-Comp A (2,2) (k=3)                                                           &   58.01                                       &   57.46  $\pm$  0.38                                                   &            0.85                                &                                   0.72         \\ \hline

XNOR2x                                                                 &  58.24                                                &   57.35  $\pm$  0.54                                                    &        0.29                                    &                                        0.3    \\ \hline
\end{tabular}
\end{table}

We observe that the proposed Hybrid-Net achieves a much superior trade-off between accuracy, energy efficiency and memory compression compared to other kinds of hybridization techniques. Moreover, in case of both ResNet-20 and ResNet-32, Hybrid-Net increases the classification accuracy by 10-11\% compared to a XNOR-Net with minimal degradation in efficiency and compression. For example, Hybrid-Net (2,2) ($\Delta=1$) based on ResNet-20 can be expected to have only 36\%  $\pm$  0.03\% of the accuracy degradation of a XNOR network with respect to a full precision network while only costing 13\% extra energy. While quantizing the entire network to 2-bit inputs and weights (Quantized (2,2)) achieves a slightly higher accuracy, we show that the our principle of increasing the bit-precision of few significant layers captures most of the increase in accuracy from an XNOR-Net to a 2-bit networks. Hybrid-Net thus consumes $14\%$ less energy and $12\%$ less memory for ResNet-20 than a 2-bit network with a performance within $2\%$ of the latter. For ResNet-32, the benefits of Hybrid-Net is even pronounced where it consumes $24\%$ less energy and $26\%$ less memory than a 2-bit network while achieving accuracy within $4\%$ of the latter. Hybrid-Net thus ensures a signficant improvement in accuracy over a binary network without making the entire network 2-bit. We also show that Hybrid-Net achieves a higher accuracy than Hybrid-Comp A networks while consuming less energy for both ResNet-20 and ResNet-32, thus demonstrating the effectiveness of the design methodology. 

\subsubsection{VGG architecture}
We further extend our analysis to VGG architectures. We considered VGG-15, which consists of 13 convolutional and 2 fully-connected layers as shown in Fig. \ref{fig:resnetvgg} (b). We kept one of the fully-connected layer binary to preserve energy-efficiency. Table \ref{results} lists the accuracy, energy efficiency and memory compression results for VGG-15 on CIFAR-100 for different networks. We consider $\Delta=3$ and $\Delta=10$ for our analysis and for each of the network configurations we use $k_b=2, 4$ for the significant layers. We observe that Hybrid-Net achieves 13\% higher accuracy than a XNOR-Net with minimal degradation in efficiency. When we make the inputs and weights of the entire network 2-bit (Quantize (2,2)), we achieve an even higher accuracy. We also show that Hybrid-Net with 2-bit layers achieve a better performance than Hybrid-Comp A for iso-efficiency in energy and memory. Similar to trends in ResNet, we observe that making the significant layers 4-bit while keeping the rest of the layers binary improves performance, however, at the cost of energy-efficiency. An entirely 2-bit network proves to be a more efficient solution. 
In summary, even for VGG architecture, we show that Hybrid-Net achieves 7.44\% higher classification accuracy compared to a XNOR network while keeping most of the layers binary.

\begin{savenotes}
\begin{table}[t]
\centering
\caption {Comparison of different networks on ImageNet}
\label{imres}
\begin{tabular}{|c|c|c|c|c|c|}
\hline
\multicolumn{6}{|c|}{\textbf{Resnet-18}}                   \\ \hline
\multicolumn{6}{|c|}{\textbf{FP Accuracy - 69.15\%}}                   \\ \hline
\multicolumn{6}{|c|}{\textbf{$E.E (XNOR)$ - 8.57}, \textbf{$M.C (XNOR)$ - 13.35}}                                                                                   \\ \hline
\multicolumn{2}{|c|}{\textit{Network Type}}     & \textit{Best Accuracy (\%)} & \textit{Mean  $\pm$  SD Accuracy (\%)} & \textit{E.E\textsubscript{Norm}}  & \textit{M.C\textsubscript{Norm}} \\ \hline
\multicolumn{2}{|c|}{XNOR}             &  50.33  & --       & 1 &   1  \\ \hline
\multicolumn{2}{|c|}{Binary-Shortcut 1} & 54.36 & 54.15  $\pm$  0.15 & 1 & 1 \\ \hline
\multicolumn{2}{|c|}{Bi-Real Net \cite{liu2018bi}} & 56.9 & -- & 1 & 1 \\ \hline
\multicolumn{2}{|c|}{Hybrid-Net (2,2) ($\Delta = 30$)}  &  60.38  & 59.75  $\pm$  0.44      & 0.96 &  0.87   \\ \hline
\multicolumn{2}{|c|}{Hybrid-Net (2,2) ($\Delta = 20$)}  &    61.95 & 61.89  $\pm$  0.04   & 0.93 &   0.8  \\ \hline
\multicolumn{2}{|l|}{Hybrid-Net (2,2) ($\Delta = 10$)}  &    62.73  & --   & 0.92 &   0.8  \\ \hline
\multicolumn{2}{|c|}{Hybrid-Net (4,4) ($\Delta = 30$)}  &   61.70 & 60.54  $\pm$  0.84  & 0.89 &   0.7  \\ \hline

\multirow{4}{*}{Quantize (2,2)} & XNOR-kbit &   64.51 & --     & \multicolumn{1}{c|}{\multirow{4}{*}{0.84}} &  \multicolumn{1}{c|}{\multirow{4}{*}{0.71}}   \\ \cline{2-3}
                                & DoReFA \cite{zhou2016dorefa} & 62.6  & --          & \multicolumn{1}{c|}{}                      & \multicolumn{1}{c|}{}                   \\ \cline{2-3}
                                & PACT \cite{choi2018pact}  & 67 & --           & \multicolumn{1}{c|}{}                      & \multicolumn{1}{c|}{}  \\ \cline{2-3}
                                & LQ-Nets \cite{zhang2018lq} & 64.9  & --          & \multicolumn{1}{c|}{}                      & \multicolumn{1}{c|}{}                   \\ \hline
\multicolumn{2}{|l|}{Hybrid \cite{prabhu2018hybrid}}    &    54.9   & --        &  0.68     &  1  \\ \hline
\multicolumn{2}{|l|}{Hybrid-Comp A (2,2) (k=4)}    &    59.47   & --        & 0.94      & 0.77    \\ \hline

\end{tabular}
\end{table}
\end{savenotes}
\subsection{Image Classification Results - ImageNet}

We evaluate the proposed Hybrid-Net design Table. \ref{imres}.   We observe that the XNOR network suffers a significant degradation in accuracy from a full-precision network. Even the Binary-Shortcut 1 network with residual connections at every layer fail to recover the classification accuracy. With improved quantization schemes and weight update mechanisms, Bi-Real Net \cite{liu2018bi} has shown a slightly higher accuracy. Compared to these binary networks, we observe that the proposed Hybrid-Net, considering both 2-bit and 4-bit weights and activations achieves upto over 10 \% higher accuracy than corresponding XNOR network. In particular, Hybrid-Net (2,2) ($\Delta=20$) can be expected to have only 38.6\%  $\pm$  0.002\% of the accuracy degradation of a XNOR network with respect to a full-precision network while only costing 7\% extra energy. Quantizing the activations and weights of the entire network to 2-bits can further increase the accuracy by 1-2\% but at the cost of a 15-20\% increase in energy consumption than Hybrid-Net. Note, that we have provided results as baselines for different input quantization algorithms, such as DoReFA-Net \cite{zhou2016dorefa}, LQ-Net \cite{zhang2018lq}, and PACT \cite{choi2018pact}, for the Quantize (2,2) network although we have used the XNOR quantization (described in Section 3.2) in this work. 
We also show Hybrid-Net (2,2) achieves upto 7.4\% higher accuracy with respect to other methods of hybridization \cite{prabhu2018hybrid}.
This work shows that increasing the bit-precision of a few significant layers can remarkably boost the performance of binary neural networks without making the entire network higher precision. Note, that using improved quantization schemes \cite{choi2018pact, liu2018bi} can potentially further increase the accuracies of the proposed Hybrid-Nets. 
\subsection{Statistical Analysis}
We have mentioned in an earlier subsection that we perform simulation for various random initializations. To  understand  the amount of variations in the results, we performed simulations for two cases: 

(a) Fixed  Optimal  Solutions: We define a binary base network with any random initialization.  We run a PCA analysis and obtain a optimal Hybrid-Net.  Next, we train the Hybrid-Net with various random initializations.

(b)Varying  Optimal Solutions: We train the base binary network with different random initializations and run PCA analysis on each case.  This provides various optimal solutions of Hybrid-Net with different combinations  of  significant  layers. Then we perform  accuracy simulations  for  corresponding random initializations.

\subsubsection{Fixed Optimal Solutions:}

We performed simulations for all cases explained in the manuscript for 5 different random initializations. For simplicity, we considered 3 random initializations for 4 networks for ImageNet. Note, the energy and memory costs depend on the network architecture and is fixed for a particular architecture. The results in Table. \ref{results} and \ref{imres} shows the variations in accuracies are within 2\% of the best accuracies. 

\subsubsection{Varying Optimal Solutions:}

We trained the base binary network for ResNet-20 on CIFAR-100 for different initializations and performed PCA analysis on each of them to obtain varying optimal solutions for different $\Delta$ values. We observe that the optimal solutions overlap quite significantly. The resulting energy efficiency and memory compression metrics also do not vary remarkably. We perform accuracy analysis for each optimal solution for their corresponding random initialization to observe the variations in results. The results are presented in Table. \ref{tabvar}. Based on the analysis, we can expect the Hybrid-Net (2,2) ($\Delta$=1) to have only 37.7  $\pm$  7E-3 \% of the accuracy degradation of a XNOR network with respect to a full-precision network with only 12 $\pm$  1.22 \% extra energy cost. Similarly, Hybrid-Net (2,2) ($\Delta$=2) is expected to have only 41  $\pm$  0.01 \% of the accuracy degradation of a XNOR network with respect to a full-precision network with only 9  $\pm$  1.4 \% extra energy cost.
\begin{table}[h!]
\centering
\caption{Analysis of effect of random initialization on varying optimal Hybrid-Net architecture (ResNet-20 on CIFAR-100)}
\label{tabvar}
\begin{tabular}{|c|c|c|c|c|c|c|}
\hline
\textit{$\Delta$}                        & \textit{Initialization} & \textit{Significant layers} & \textit{Best Accuracy (\%)} & \textit{Mean   $\pm$  SD Accuracy (\%)} & \textit{$E.E_{Norm}$} & \textit{$M.C_{Norm}$} \\ \hline
\multicolumn{1}{|c|}{\multirow{5}{*}{1}} & 1                       & 8,9,10,14,15,16,18                    &    62.84             &     62.46 $\pm$ 0.23  & 0.88                  & 0.77                  \\ \cline{2-7} 
\multicolumn{1}{|c|}{}                   & 2                       & 5,8,9,14,15,16                        &      62.14      &      61.59 $\pm$ 0.29       & 0.89                  & 0.82                  \\ \cline{2-7} 
\multicolumn{1}{|c|}{}                   & 3                       &         7,8,9,14,15,16                              &   62.17     &    61.85 $\pm$ 0.19             & 0.89                  & 0.82                  \\ \cline{2-7} 
\multicolumn{1}{|c|}{}                   & 4                       &           2,8,9,10,14,15,16,18                              &     63.4        &   63.15 $\pm$ 0.13        & 0.86                  & 0.77                  \\ \cline{2-7} 
\hline
\multirow{5}{*}{2}                       & 1                       &            8,9,14,15                         &         61.49       &    60.72 $\pm$ 0.52     & 0.93                  & 0.88                  \\ \cline{2-7} 
                                         & 2                       &     8,9,14,15,16
                                  &         61.29    &   61.48 $\pm$ 0.29         & 0.91                  & 0.83                  \\ \cline{2-7} 
                                         & 3                       &  6,8,9,14,15,16
                                     &        61.70    &   61.82 $\pm$ 0.53          & 0.89                  & 0.82                  \\ \cline{2-7} 
                                         & 4                       &    8,9,14,15,16
                                   &    61.87         &    61.48 $\pm$ 0.29        & 0.91                  & 0.83                  \\ \cline{2-7} 
 \hline
\end{tabular}
\end{table}

Following this analysis, we can conclude that variation in classification accuracy due to different random initializations vary within a range of 2\% for all cases considered. Further, it is true that random initializations lead to varying optimal solutions which would mean the PCA step needs to be performed for every network initialization and dataset. For edge applications, the network can be tuned for a target application during initial stages of deployment.

\subsection{Optimality Studies}
\begin{figure*}[h!]
	\centering
	\includegraphics[width=5.5in, keepaspectratio]{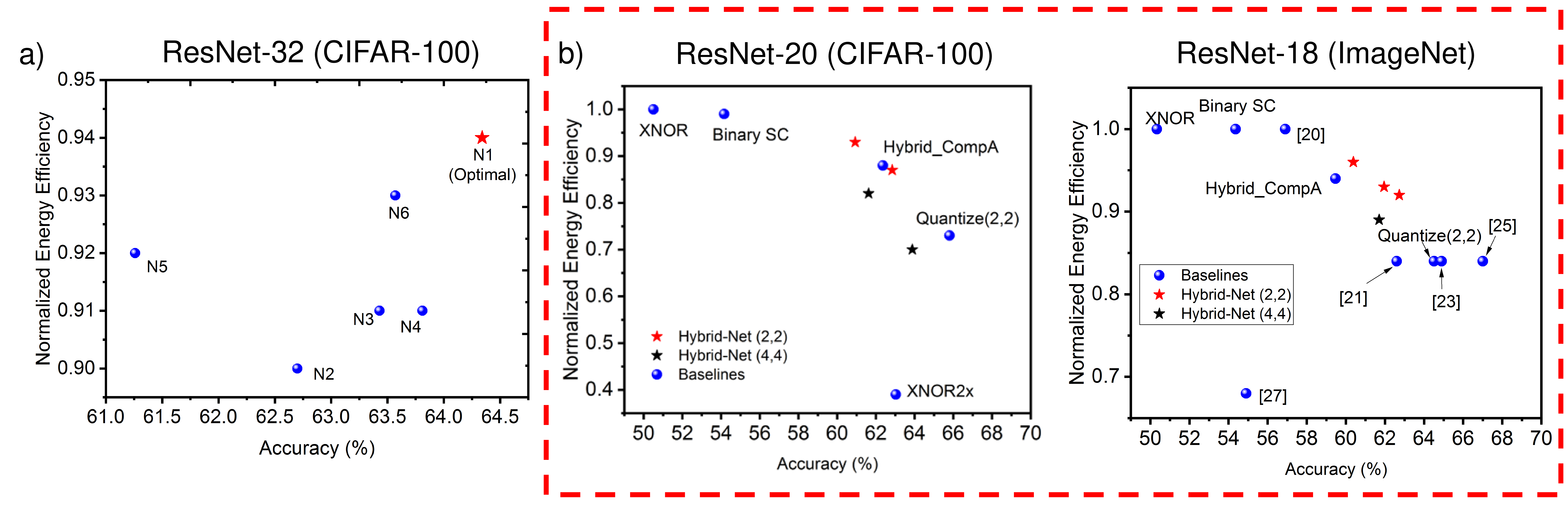}

	\caption{Illustration of energy-accuracy optimality of Hybrid-Net. a) Accuracy v/s Energy efficiency plot showing that PCA-driven Hybrid network design achieves more optimal tradeoffs than randomly chosen layers.. b) Normalized Energy Efficiency v/s Accuracy plot showing optimality boundaries for the considered network configurations. It shows that Hybrid-Net (2,2) networks lie right on the optimal boundary among the networks considered. Note, that PACT involves a more advanced quantization algorithm than other networks.}
	\label{fig:par2}

\end{figure*}

The optimality of the proposed Hybrid-Net configurations can be understood through two visualizations. First, we compare the network designed through the proposed PCA-driven methodology with arbitrarily defined network architectures with randomly chosen layers as $k_b$-bit precision. We have performed this analysis on ResNet-32 for CIFAR-100 dataset. For identical comparisons to the proposed network Hybrid-Net(2,2) ($\Delta=1$), we have defined networks with randomly chosen 6 or 7 layers with $k_b$-bit precision. Table. \ref{tab1} shows the networks. The numbers within the bracket signifies the layers with $k_b$-bit precision. The rest of the layers in this network is binary. Fig. \ref{fig:par2} (a) shows the resulting plot. 

\begin{table}[h!]

\caption{Network Configurations with randomly chosen layers as $k_b$-bit precision}
\label{tab1}
\centering
\begin{tabular}{|c|c|c|c|c|}
\hline
\textit{Network index} & \textit{Network Configurations}  & \textit{Best Accuracy (\%)}       & \textit{Mean  $\pm$  SD Accuracy (\%)} & \textit{Energy}  \\ \hline N1 &
Hybrid-Net (2,2) (Delta=4)              & 64.34       &   63.75 $\pm$ 0.39   &       0.94                           \\ \hline
N2 & Hybrid-Net (25, 26, 27, 28, 29, 30, 31) & 62.70     &  62.62 $\pm$ 0.09      &      0.90                            \\ \hline
N3 & Hybrid-Net (2, 11, 12, 20, 21, 23, 29 ) & 63.43    &    63.00 $\pm$ 0.37     &   0.91                              \\ \hline
N4 & Hybrid-Net (12, 17, 18, 20, 24, 25, 26) &   63.81   &   63.46 $\pm$ 0.21         &  0.91                               \\ \hline
N5 & Hybrid-Net (2, 5, 6, 7, 20, 28)         &      61.26  &    61.04 $\pm$ 0.24       &    0.92                          \\ \hline
N6 & Hybrid-Net (2, 17, 22, 25, 28, 30)      &   63.57   &    63.43 $\pm$ 0.12         &       0.93                          \\ \hline
\end{tabular}
\end{table}

We observe that hybrid network configurations with similar energy efficiencies lead to different accuracies. This gives the intuition that some layers might be more significant than others. This is the premise our technique is based on where we provide a methodology to identify those significant layers. We observe that PCA-driven methodology of designing Hybrid-Net achieves a better energy-accuracy tradeoff than networks with randomly chosen layers as $k_b$-bit precision. 

Another visualization of optimality would be a comparison of the networks with varying $\Delta$ and other baselines considered in this work. Here, we plot the best accuracies obtained for different configurations of ResNet-20 on CIFAR-100 and ResNet-18 on ImageNet described in Table. \ref{results} and Table. \ref{imres}. Fig. \ref{fig:par2} (b) shows such a plot. 

Note, that its difficult to comment on Pareto frontier without doing an exhaustive search of networks which can prove to be quite time expensive. In this work, we have focused on improving the accuracy of extremely quantized neural networks without a significant degradation in energy efficiency. Our technique proposes a methodology to design hybrid networks through a PCA-driven significance analysis, which achieves ~10\% higher accuracy with less than 6-10\% increase in energy consumption. Out of the network configurations considered, Hybrid-Net (2,2) configurations shows that the proposed design principle can lead us to optimality for both CIFAR-100 and ImageNet. The absolute Pareto optimality is difficult to gauge without performing an exhaustive search. 
\subsection{Discussion}
The proposed Hybrid-Net design uses PCA-driven hybridization of extremely quantized neural networks, resulting in significant improvements and observations as listed. One key contribution of the proposed methodology is that we can design hybrid networks without any iterations. It does not require an iterative design space exploration to identify optimal networks. Moreover, this methodology shows that increasing the bit-precision of only the significant layers in a binary network achieves performance close to that of a network that is entirely composed of layers with higher bit-precision weights and activations. Intuitively, a 2-bit network performs much better than a binary network. However, our analysis shows that it is not necessary to make the weights and activations of the entire network 2-bit. Hybrid-Net achieves more than $\sim 10\%$ improvement over a XNOR network, which is a fully binary network except the first and final layers, by increasing the bit-precision of less than half of the entire network. In fact, for deeper networks, like ResNet-34, this improvement is achieved with only $\sim 6\%$ increase in energy consumption from a XNOR-Net \cite{rastegari2016xnor}. Thus, Hybrid-Net goes a long way in reaching close to high-precision accuracies with networks which are mostly binary and attain comparable energy-efficiency and memory compression to binary networks such as XNOR-Net \cite{rastegari2016xnor} and BNN \cite{courbariaux2016binarized}. Moreover, this methodology can be extended to any network where making significant layers of the network $k_{b2}$-bit while keeping the rest of the network $k_{b1}$-bit ($k_{b2}>k_{b1}$), can potentially produce comparable performance with enhanced energy-efficiency than an entirely $k_{b2}$-bit network. 

Secondly, the performance of Hybrid-Net is subject to the nature of the plots obtained from PCA on the binary version of the networks. For example, for ResNet architectures (Fig. \ref{fig:pca} (a), (b) and (c) for CIFAR-100 and Fig. \ref{fig:pca} (d) for ImageNet), we observe the number of significant components increase for the layers which are adjacent to the ones where the number of output channels increase and then decrease for the later layers which have the same number of output channels. It can be said that the later layers are not adding to the linear separability of the data and binarizing them preserve the accuracy as observed. Or in other words, the significant layers identified using our proposed methodology contribute remarkably higher to the linear separability than the other layers. This is reflected in the results where we show the performance difference between Hybrid-Net and a 2-bit network is less than $4\%$. However, for VGG architectures, we observe that the PCA plot remains fairly flat, which means that the identified significant layers are not remarkably different in their contribution towards linear separability of the data in comparison to the other layers. This is reflected in the performance difference ($> 6\%$) of Hybrid-Net from a 2-bit network for a VGG-15 network. 

Thirdly, we observe that increasing the bit-precision of the weights and activations of the significant layers to 4-bits while keeping the rest of the layers binary is not the most energy-efficient way of improving accuracy of a network. An entire 2-bit network proves to be more energy-efficient while performing better. It may be because the loss due to binarization can not be significantly recovered by increasing the bit-precision much higher than binary, while keeping most of the layers binary. Thus, the proposed methodology performs best when the precision of the significant layers is close to the base precision of the network (binary in our case).

Finally, the precision of the first and final layers is of utmost important for extremely quantized neural networks. To study the influence of quantization on these layers, we performed experiments on our Hybrid-Net designs by making the weights of the final layer binary while having full-precision inputs as suggested in [27]. The results for ResNet-32 for CIFAR-100 for the configuration Hybrid-Net(2,2) ($\Delta$=4) are listed below:

\begin{table}[h!]
\centering
\caption{Analysis of quantization of first and last layers in Hybrid-Net (2,2) (Delta=4) on ResNet-32 for CIFAR-100}
\begin{tabular}{|l|c|c|c|c|}
\hline
\textit{Last Layer Configuration}             & \textit{Best Accuracy (\%)} & \textit{Mean  $\pm$  SD Accuracy (\%)} & \textit{Energy Consumption} & \textit{Memory}\\ \hline
Full-Precision                                & 64.34    &   63.75 $\pm$ 0.39      & 1  & 1                         \\ \hline
Binary weights and activations                & 56.93     &    56.83 $\pm$ 0.09    & 0.98 & 0.75                       \\ \hline
Binary weights and full-precision activations & 61.07      &    60.36 $\pm$ 0.44   & 0.98  & 0.75                      \\ \hline
2-bit weights and activations  & 62.41      &   62.35 $\pm$ 0.05    & 0.98      & 0.76                  \\ \hline
\textit{First Layer Configuration}            & \textit{Best Accuracy (\%)} & \textit{Mean  $\pm$  SD Accuracy (\%)}  & \textit{Energy Consumption}   & \textit{Memory}          \\ \hline
Binary weights and activations                &     44.79      &   44.16 $\pm$ 0.74     & 0.85 &  0.98                        \\ \hline
Binary weights and full-precision activations &       59.94     &  59.29 $\pm$ 0.57     & 0.93   & 0.98                       \\ \hline
2-bit weights and activations  &             60.87  &  60.46 $\pm$ 0.25  & 0.85         &  0.98               \\ \hline

\end{tabular}
\end{table}

\textit{Energy Consumption:}
The energy consumption reduction after quantizing the last layer of the neural network is only 2\% for CIFAR-100. On the other hand, quantizing the last layer leads to at least 2\% degradation in accuracy (when the last layer is 2-bit precision). The first layer consumes more energy due to larger output map size. We performed this analysis by quantizing the first layer as well. We observe that although the energy consumption reduces by 15\%, the accuracy also degrades by 3.5\%. 

\textit{Memory:}
The memory requirements of the last layer is a significant aspect. We observe that quantizing the last layer can lead to close to 25\% lower memory. In case of ImageNet, this will be even more significant and the memory reduction can be upto 2x. However, fully binarizing the last layer leads to significant accuracy degradation. Thus, keeping them higher precision (2-bit or 4-bit) will lead to better accuracy-memory tradeoffs. 

In this work, we have considered the quantization scheme, explored in \cite{rastegari2016xnor}. Since then, there has been a plethora of works focused on improving quantization for both inputs and weights \cite{zhou2016dorefa, choi2018pact}. Hybrid-Net focuses on improving the performance of binary neural networks through mixed-precision network design and we believe the improved quantization schemes should further increase the accuracy of both Hybrid-Nets and entirely 2-bit or 4-bit networks. 


The humongous computing power and memory requirements of deep networks stand in the way of ubiquitous use of AI for performing on-chip analytics in low-power edge devices. The significant energy efficiency offered by the compressed hybrid networks increases the viability of using AI, powered by deep neural networks, in edge devices. With the proliferation of connected devices in the IOT environment, AI-enabled edge computing can reduce the communication overhead of cloud computing and augment the functionalities of the devices beyond primitive tasks such as sensing, transmission and reception to in-situ processing.

\section{Conclusion}
Binary neural networks offer significant energy-efficiency and memory compression compared to full-precision networks. In this work, we propose a one-shot methodology for designing mixed-precision, hybrid networks with binary and higher bit-precision inputs and weights to improve the performance of extremely quantized neural networks in terms of classification accuracy while still achieving significant energy efficiency and memory compression. The proposed methodology uses PCA to identify significant layers in a binary network which transform the input data such that the output feature space require more significant dimensions to explain variance in data. PCA is usually exploited to perform layer-wise dimensional reduction. We use PCA in an opposite manner in order to determine which layers cause the number of signficant dimensions to increase across input and output. Next, we increase the bit-precision of the weights and activations of the significant layers and keeping that of the other layers binary. The proposed Hybrid-Net achieves more than $\sim 10\%$ improvement over XNOR networks for ResNet and VGG network architectures on CIFAR-100 and ImageNet with only $\sim 6-10\%$ increase in energy consumption, thus ensuring more than $15-20x$ reduction in energy consumption and memory compression from full-precision networks. Memory compression along with the close match to high-precision accuracies offered by the proposed mixed-precision network design using layer-wise information allows us to explore interesting possibilities in the realm of hardware-software co-design. This work thus proposes an effective, one-shot methodology for designing hybrid, compressed neural networks and potentially paves the way toward using energy-efficient hybrid networks for AI-based on-chip analytics in low-power edge devices with accuracy close to full-precision networks.

\section{Methods}
\subsection{Energy Efficiency and Memory calculations}
\subsubsection{\textbf{Energy Efficiency}}
The primary model-dependent metrics that affect the energy consumption of classification task are the energies consumed by the computations (multiply-and-accumulate or MAC operations) and memory accesses in our calculations for energy efficiency. We exclude energy consumed due to data flow and instruction flow in the architecture. For a convolutional layer, there are $I$ input channels and $O$ output channels. Let the size of the input be $N\times N$, size of the kernel be $k \times k$ and size of the output be $M\times M$. Thus, in Table \ref{kbops} we present the number of memory-accesses $N_{M-FP}$ and computations $N_{C-FP}$ for standard full-precision (FP) networks:

The number of binary memory accesses ($N_{Mi}$) and computations ($N_{Ci}$) in a binary layer is same as the corresponding number in full-precision layer of equivalent dimensions. As explained in Eq. 1, we consider additional full-precision memory accesses and computations for parameter $\alpha$, where $\alpha$ is the scaling factor for each filter bank in a convolutional layer. Number of accesses for $\alpha$ is equal to the number of output maps, $O$. Number of full-precision computations are $M^2\times O$.  Table \ref{kbops} lists the number of k-bit and full-precision memory access and computations of any layer. 
\begin{table}[h!]
\centering
	\caption{Number of operations in a $k_b$-bit layer}

	\label{kbops}
	\begin{tabular}{|l|l|l|}
		\hline
			\multicolumn{3}{|c|}{Operations in neural networks}  \\ \hline\hline
		Operation          & \multicolumn{2}{|c|}{Number of Operations}               \\ \hline\hline
		Input Read         & \multicolumn{2}{|c|}{$N^2\times I$}                      \\ \hline
		Weight Read        & \multicolumn{2}{|c|}{$k^2\times I\times O$}           \\ \hline
		Computations (MAC) & \multicolumn{2}{|c|}{$M^2 \times I \times k^2 \times O$} \\ \hline
		Memory Write       & \multicolumn{2}{|c|}{$M^2\times O$}                      \\ \hline \hline
		
			\multicolumn{3}{|c|}{Number of operations of $k_b$-bit layer} \\ \hline \hline
		Operation          & Term & Number of Operations               \\ \hline\hline
		k-bit Memory Access  &  $N_{A-k}$     & $N^2\times I$+ $k^2\times I\times O$                     \\ \hline
		k-bit Computations (MAC)& $N_{C-k}$       & $M^2 \times I \times k^2 \times O$              \\ \hline
		FP Memory Access  & $N_{A-F}$ &$O$ \\ \hline
		FP Computations   & $N_{C-F}$   & $M^2\times O$                      \\ \hline \hline
		\multicolumn{3}{|c|}{Energy Consumption Chart} \\ \hline \hline
		Operation         & Term & Energy (pJ) \\ \hline
k-b Memory Access& $E_{A-k}$ & 2.5$k$          \\ \hline
32-b MULT FP      & $E_{M-F}$ & 3.7         \\ \hline
32-b MULT INT     & $E_{M-I}$ & 3.1         \\ \hline
32-b ADD FP       & $E_{AD-F}$ & 0.9         \\ \hline
32-b ADD INT      & $E_{AD-I}$ & 0.1         \\ \hline
k-bit MAC INT        & $E_{C-kI}$ & ((3.1*$k$)/32+0.1) \\ \hline
k-bit MAC FP       & $E_{C-kF}$ & 4.6 \\ \hline
	\end{tabular}
\end{table}
We calculated the energy consumption from projections for 45 nm CMOS technology \cite{keckler2011gpus,han2015learning}. Considering 32-bit representation as full-precision, the energy consumption for both binary and 32-bit memory accesses and computations are shown in Table. \ref{kbops}.

Then, energy consumed by any layer with k-bit weights and activations is given by 
\begin{equation}
    E_i = N_{A-F}E_{A-32}+N_{A-k}E_{A-k}+N_{C-F}E_{C-kF}+N_{C-k}E_{C-kI}
\end{equation}

 Note, this calculation is a rather conservative estimate which does not take into account other hardware architectural aspects such as input-sharing or weight-sharing. However, our approach concerns with modifications of network architecture and we compare the ratios of energy consumption. These aspects of the hardware architecture affect all the networks equally and hence can be taken out of consideration. Further, FP MAC operations can be optimized for lower energy consumptions. In our calculations, we have bluntly taken it as the sum of a 32-b FP Multiply and 32-b FP Add operations. These optimizations are catered towards FP networks, and reduce the FP energy consumption. This, in turn, will reduce the energy efficiency of the binary and hybrid networks. In this work, we are focused on comparing different kinds of binary and hybrid network, and hence, this assumption of FP MAC energy is not going to affect the analysis.

\subsubsection{\textbf{Memory Compression}}
The memory required for any network is given by product of the total number of weights in the network multiplied by the precision of the weights. The number of weights in any layer is given by: 

\begin{equation}
    N_{w-i} = I_i\times O_i\times k^2
\end{equation}
considering usual notations describer earlier. Thus, the total memory requirements can be simply written as $M_i = \sum_i N_{w-i}*k_{b-i}$ where $k_{b-i}$ is the precision of weights in the $i^{th}$ layer. We can estimate memory compression (M.C) with respect to a full-precision network and normalize it with respect to an XNOR-Net network which is an entirely binary network except the first and final layer. 


Note that the assumption for the energy and storage calculations for binary layers hold for custom hardware capable of handling fixed-point binary representations of data, thus leveraging the benefits offered by quantized networks. 

\subsection{Implementation and Software Package Details}
\subsubsection{Data Availability Statement}
All datasets used in this work are publicly available: CIFAR-100 \cite{krizhevsky2009learning} and ImageNet \cite{deng2009imagenet}.

\subsubsection{Code Availability Statement}
Publicly available tools, Python and PyTorch, were used to perform the experiments. The custom codes for the work are available at : https://github.com/ichakra2/pca-hybrid.
\section*{Competing Interests}
The authors declare no competing interests.
\section*{Acknowledgement}
This work was supported in part by the Center for Brain-inspired Computing Enabling Autonomous Intelligence (C-BRIC), one of six centers in JUMP, a Semiconductor Research Corporation (SRC) program sponsored by DARPA, in part by the National Science Foundation, in part by Intel, in part by the ONR-MURI program and in part by the Vannevar Bush Faculty Fellowship.
\section*{Contributions}
I.C and K.R conceived the idea. I.C, D.R and I.G developed the simulation framework. I.C carried out all experiments. I.C and A.A developed the energy and memory analysis framework. I.C, D.R, I.G and K.R analyzed the results. I.C, D.R and I.G wrote the paper.

\bibliography{sample}

\end{document}